\documentclass[11pt]{article}

\usepackage[preprint]{acl}

\usepackage{times}
\usepackage{latexsym}

\usepackage{paralist}
\usepackage{natbib}
\usepackage{graphicx}
\usepackage{booktabs}
\usepackage{array}
\usepackage{booktabs}
\usepackage{multirow}
\usepackage[accsupp]{axessibility}  

\usepackage{arydshln}
\usepackage{makecell}
\usepackage{multirow}

\usepackage[T1]{fontenc}

\usepackage[utf8]{inputenc}

\usepackage{microtype}

\usepackage{inconsolata}

\usepackage{graphicx}

\title{TRACE: Training-aware Robustness and Challenge Evaluation for Text-Vision Embedding Models}

\author{
 \textbf{Pume Tuchinda\textsuperscript{1}},
 \textbf{Tinnakit Udsa\textsuperscript{2$\dagger$}},
 \textbf{Peerat Limkonchotiwat\textsuperscript{1}}
 \\
 \textsuperscript{1}AI Singapore,
 \textsuperscript{2}School of Information Science and Technology VISTEC
 \\
 \texttt{\{pume,peerat\}@aisingapore.org},
 \texttt{tinnakit.u\_s24@vistec.ac.th}
 \\
\href{https://huggingface.co/collections/airesearch/trace-training-aware-robustness-and-challenge-evaluation-fo}{\includegraphics[height=1em]{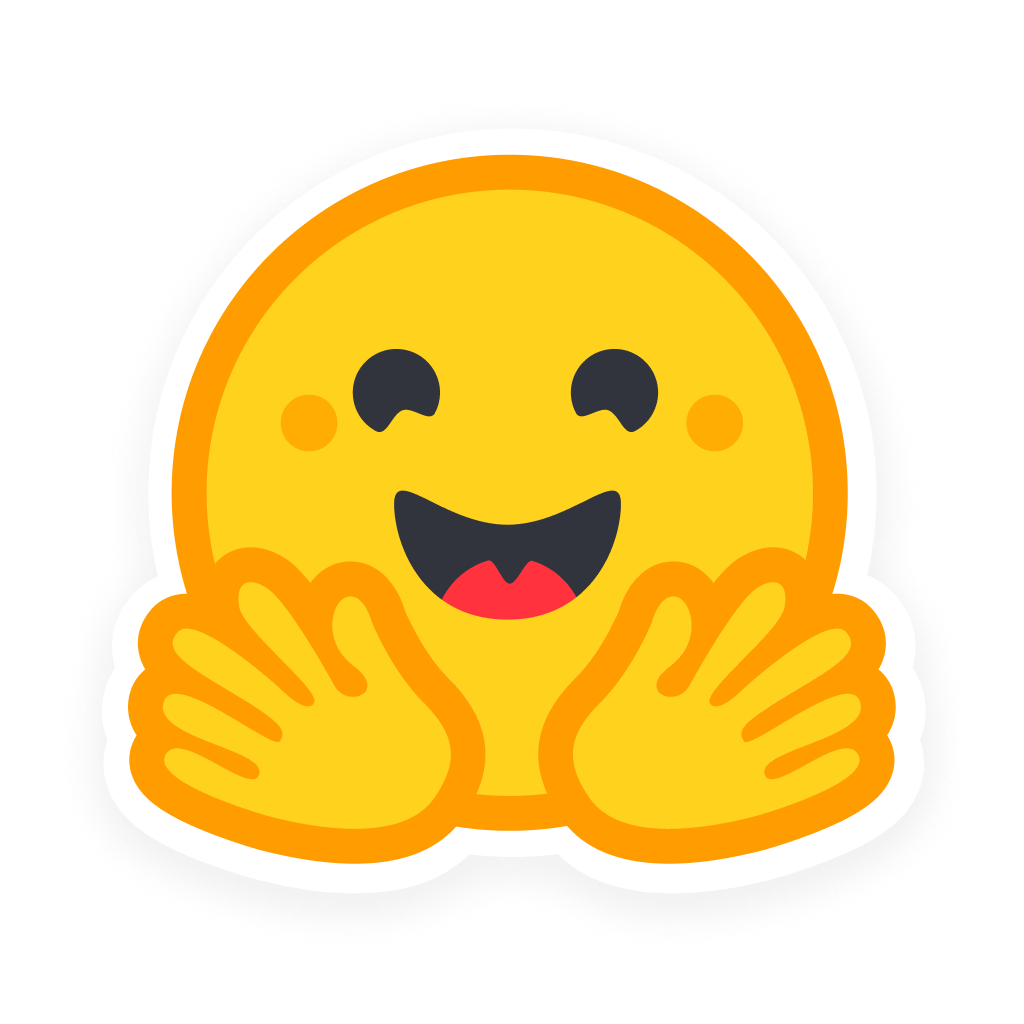}\ Hugging Face}
}

\begin{document}
\maketitle

\begingroup
\renewcommand\thefootnote{$\dagger$}
\footnotetext{Work was conducted while Tinnakit Udsa was a visiting scholar at AI Singapore}
\endgroup

\begin{abstract}

Existing benchmarks for text-vision embedding models report aggregate scores that obscure whether performance comes from general capabilities, task-specific adaptation, or robustness to challenging tasks.
We introduce \textbf{TRACE}, a Training-aware Robustness and Challenge Evaluation framework that reorganizes 168 tasks from various sources of datasets into three categories: \emph{General}, \emph{Adaptability}, and \emph{Challenge}. 
Through controlled experiments on two text-vision embedding models, we show that combining general and adaptation data improves Adaptability performance by 7.67 and 7.09 points, respectively, while largely preserving General performance. 
However, Challenge performance remains below 5.0, with limited improvements from additional training data. Further analysis identifies multilingual and visual-document tasks as major weaknesses. 
Our findings demonstrate that data adaptation alone is insufficient to address these challenges, highlighting the need for methodological advances beyond data-centric improvements.

\end{abstract}

\section{Introduction}
\label{sec:introduction}

Text-vision embedding models enable applications such as image-text retrieval, visual question answering, clustering, and multimodal search~\cite{radford2021learning, zhai2023sigmoid}. 
To evaluate these models in real applications, recent benchmarks, such as MMEB~\cite{mmeb} and MIEB~\cite{mieb}, provide broad evaluation suites for multimodal embedding models, covering diverse task types from applications.
%
%
These benchmarks have been used to evaluate how much text-vision embedding models align with real-world scenarios~\cite{meng2025vlm2vec}; higher performance means higher robustness in applications.

That said, these benchmarks often combine datasets spanning diverse tasks and difficulty levels, treating each dataset and task equally.
%
As a result, model development tends to chase higher aggregate scores, offering little insight into where current models fall short or how they can be improved.
%
Such aggregate scores overlook task difficulty and are hard to interpret.
%
Moreover, they do not account for the relationship between training and evaluation data.
A model may perform well because the task is broadly solvable from large-scale pretraining, related in-domain training data is available, or the evaluation set closely resembles the training distribution.

In this work, we propose \emph{TRACE}, a \textbf{T}raining-aware \textbf{R}obustness \textbf{A}nd \textbf{C}hallenge \textbf{E}valuation framework for text-vision embeddings.
TRACE organizes training and evaluation data according to the difficulty of the benchmark to improve the interpretability of the results. 
In particular, this property enables us to assess the existing capabilities of current and newly developed algorithms/models and identify areas for further improvement.
TRACE separates evaluation into three groups, each with distinct goals:
(i) \textbf{General} measures tasks that can use common text-vision training data, i.e., as CC12M~\cite{cc12m}, to improve the performance without relying on a specific training dataset. 
The tasks in this group enable the measurement of the general capability of models.
(ii) \textbf{Adaptability} measures the adaptation of models when models adapt to a new task/domain, e.g., training on MIMIC-CXR~\cite{johnson2019mimic} to improve the skill on medical/legal domains. 
(iii) \textbf{Challenge} group comprises evaluation datasets on which all models perform poorly to evaluate the robustness of models in the out-of-distribution case.
%

Our experiments show that TRACE reveals performance differences obscured by the aggregate scores of MIEB and MMEB. 
Combining General and Adaptability data improves Adaptability by 7.67 and 7.09 points for CLIP-KD and TinyCLIP, respectively, while largely preserving General capabilities. 
At the task level, STS and VQA/Choice are the most challenging in General and Adaptability, respectively, with improvements mainly driven by Classification. 
However, Challenge performance remains consistently low (below 5.0 on average), with negligible gains from additional data, suggesting that data-centric improvements alone are insufficient to address these challenges.

We summarize the contributions of this paper as follows. \textbf{Data:} We propose a training setup that separates general training data from benchmark-related in-domain data, enabling clearer analysis of generalization and adaptation. \textbf{Evaluation:} We introduce TRACE, a training-aware evaluation framework that interprets benchmark results according to the relationship between training and evaluation data. \textbf{Study:} We conduct a controlled study of text-vision embedding models to identify which tasks are solvable by general training, which require in-domain adaptation, and which remain challenging.

\section{TRACE}
\label{sec:method}

To formulate a benchmark that can interpret model performance, we conduct \emph{TRACE} with two main components: (i) a controlled training setup and (ii) a training-aware evaluation protocol.
Given existing text-vision embedding benchmarks~\cite{mieb,mmeb,mmeb_v2}, TRACE groups tasks into \textbf{General}, \textbf{Adaptability}, and \textbf{Challenge}, then retrains representative small text-vision embedding models under controlled data settings.



\begin{table}[h!]
\vspace{-3mm}
\centering
\small
\setlength{\tabcolsep}{2pt}
\scalebox{0.8}{
\begin{tabular}{lcccc}
\toprule
\textbf{Benchmark} & \textbf{\#Tasks} &  \textbf{\#Data} & \textbf{\#Lang.}  & \textbf{\makecell{Difficulty\\Levels?}} \\
\midrule
UnED~\cite{uned}   & 8 & 8 & 1  & \texttimes \\
M-BEIR~\cite{m_beir} &  8  &   10     & 1 &  \texttimes      \\
MMEB~\cite{mmeb}   & 36   &  31    &  1 & \texttimes         \\
MMEB v2~\cite{mmeb_v2}$^\dagger$   & 63   &     48    & 4 &   \texttimes         \\
MIEB~\cite{mieb}   & 130   &   79   &  38 & \texttimes         \\
\midrule
TRACE  &  171  & 102   & 38 & \checkmark \\
\bottomrule
\end{tabular}}
\vspace{-2mm}
\caption{Comparison of TRACE with prior text-vision embedding benchmarks. TRACE covers more tasks and datasets, and adds interpretability to reveal model strengths and weaknesses. \#Data counts unique source datasets; a dataset evaluated in several directions or task types (e.g.\ MSCOCO i2t and t2i) is counted once. \#Lang. counts unique languages in benchmark; $^\dagger$MMEB v2 is reported without its 18 video tasks.}
\vspace{-5mm}
\end{table}

\subsection{How Do We Collect Datasets?}

We construct our evaluation suite from MMEB v1~\cite{mmeb} and v2~\cite{mmeb_v2}, an evaluation benchmark for multimodal embedding models across diverse task families, and 2) MIEB~\cite{mieb}, a broad evaluation suite for image and text-image embedding models across many tasks and languages.
Since our focus is on text-image embedding models, we exclude video tasks from MMEB v2.
When the same task appears in both MMEB and MIEB, we remove the duplicate and keep the MIEB version.
We combine and get 171 tasks from 102 datasets: 108 MIEB tasks (77 datasets), 36 MMEB image tasks (31 datasets) and 27 MMEB visdoc tasks (20 datasets).
We organize the benchmark at the task level, since the same dataset may support different task formats.
For each dataset, we keep the original evaluation format and metric.
Together, these benchmarks provide a comprehensive view of embedding quality.


\begin{table}[h!]
\vspace{-2mm}
\centering
\setlength{\tabcolsep}{2pt}
\scalebox{0.75}{
\begin{tabular}{l c c c l}
\toprule
\textbf{Group} & \textbf{\#Tasks} & \textbf{\#Datasets} & \textbf{\#Train Samples} & \textbf{Purpose?} \\ \midrule
General & 48 & 31 & - & \footnotesize{Basic capability} \\
Adaptability & 72 & 43 & 5.5M & \footnotesize{Adaptation} \\
Challenge & 51 & 33 & - & \footnotesize{Robustness} \\
\bottomrule
\end{tabular}}
\vspace{-1mm}
\caption{
The statistics of TRACE and the goal of each evaluation group.
}
\label{tab:trace_groups}
\vspace{-7mm}
\end{table}

\subsection{How Do We Group Each Task?} \label{subsec:grouping}

TRACE groups tasks using both dataset-level score and controlled model performances.
%
%
For the controlled models, we use two small, famous CLIP models, namely CLIP-KD~\cite{clipkd} and TinyCLIP~\cite{tinyclip} from the original source.
%
%
%
%
%
To gather the dataset-level score, we evaluate the task score using controlled models, where we average the score of those models for each task.
Let $\tau$ be a fixed performance threshold.
We use $\tau$ to group the task into categories as follows:

\begin{compactitem}[\hspace{2mm}•]
    \item First, if tasks provided the training data, we group these tasks into  \textbf{Adaptability}. In total, we gather training data from MMEB, both v1 and v2, and MIEB, for 5.5M samples. 

    \item Second, if tasks whose average score is greater than or equal to $\tau$ and did not provide any training data, we group these tasks into \textbf{General}.

    \item Lastly, if tasks whose average score is less than $\tau$ and did not provide any training data, we group these tasks into \textbf{Challenge}.

\end{compactitem}

%




\begin{figure}[h!]
\vspace{-1mm}
\centering
\includegraphics[width=\linewidth]{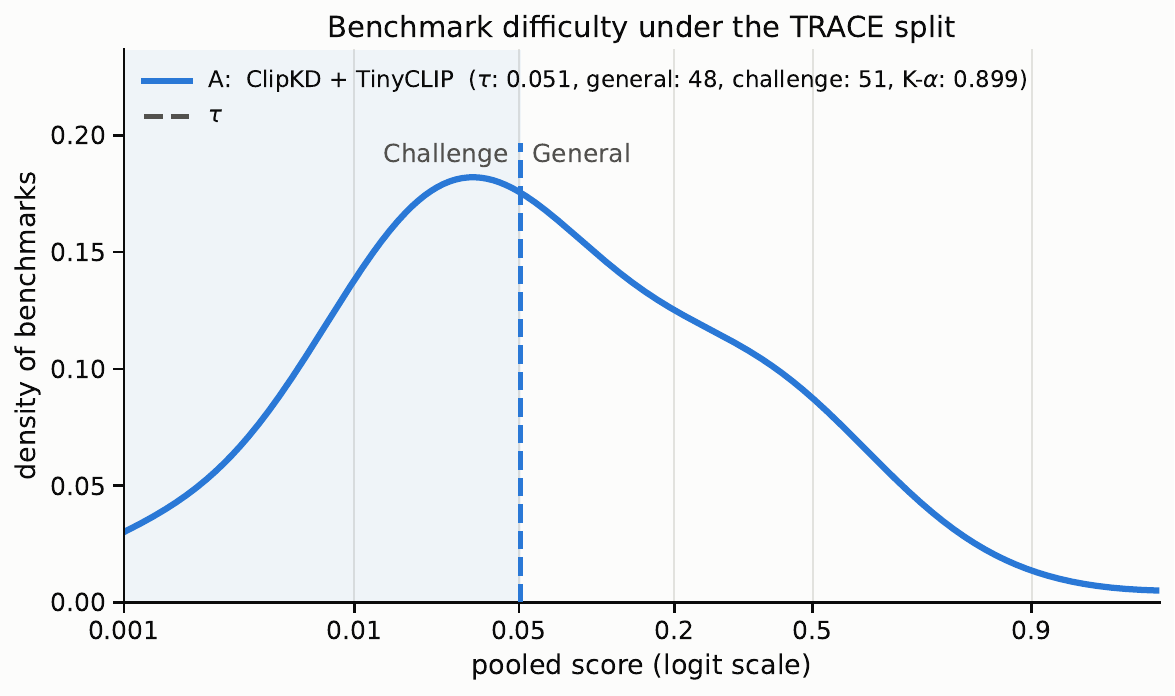}
\vspace{-8mm}
\caption{Kernel density estimate of the pooled CLIP-KD + TinyCLIP score over the 99 non-adaptable benchmarks (logit scale). $\tau$ is the median of the KDE mixture and splits them into General (48) and Challenge (51); K-$\alpha$ is Krippendorff's alpha between the pool's models on that split.}
\vspace{-6mm}
\label{fig:split_kde_A}
\end{figure}

To set $\tau$, we use a kernel density estimate of the task score. Treating each controlled model as an annotator, every task without training data gets a Gaussian density over its annotators' scores with a shared bandwidth, and $\tau$ is the median of the mixture of these densities, which gives $\tau = 0.050$ (Figure~\ref{fig:split_kde_A}).
%
%
As shown in Table~\ref{tab:trace_groups}, TRACE contains 48 General tasks, 72 Adaptability tasks, and 51 Challenge tasks.
The full list of datasets and their groups are in Table~\ref{tab:group_members}
Additional studies for choice of $\tau$ is in Appendix~\ref{sec:tau}.

\begin{table*}[h!]
\centering
\setlength{\tabcolsep}{3.5pt}
\small
\begin{tabular}{l l | c | c c c | c c}
\toprule
\textbf{Model} & \textbf{Training} & \textbf{Overall} & \textbf{General} & \textbf{Adapt.} & \textbf{Challenge} & \textbf{MIEB} & \textbf{MMEB} \\
\midrule
Teacher CLIP & - & 39.23 & 52.52 & 49.09 & 16.08 & 44.65 & 32.60 \\
\hdashline
\multirow{4}{*}{CLIP-KD} & Original & 24.17 & 35.44 & 34.86 & 2.21 & 30.27 & 16.73 \\
 & General & 24.03 $\pm$ 0.21 & 35.19 $\pm$ 0.30 & 34.49 $\pm$ 0.23 & 2.41 $\pm$ 0.13 & 30.17 $\pm$ 0.24 & 16.46 $\pm$ 0.19 \\
 & Gen.+Adapt. & 26.67 $\pm$ 0.13 & 35.30 $\pm$ 0.39 & 42.06 $\pm$ 0.11 & 2.66 $\pm$ 0.12 & 34.22 $\pm$ 0.17 & 18.45 $\pm$ 0.05 \\
 & Adapt. & 11.58 $\pm$ 1.37 & 19.50 $\pm$ 1.65 & 14.49 $\pm$ 2.25 & 0.74 $\pm$ 0.20 & 16.49 $\pm$ 1.47 & 3.76 $\pm$ 1.46 \\
\hline
Teacher CLIP & - & 40.85 & 52.73 & 51.09 & 18.74 & 46.04 & 34.80 \\
\hdashline
\multirow{4}{*}{TinyCLIP} & Original & 22.50 & 32.84 & 32.86 & 1.81 & 28.33 & 15.47 \\
 & General & 21.06 $\pm$ 0.19 & 31.82 $\pm$ 0.70 & 29.82 $\pm$ 0.21 & 1.55 $\pm$ 0.12 & 26.86 $\pm$ 0.21 & 13.54 $\pm$ 0.07 \\
 & Gen.+Adapt. & 23.89 $\pm$ 0.03 & 32.83 $\pm$ 0.10 & 36.83 $\pm$ 0.10 & 2.01 $\pm$ 0.04 & 31.12 $\pm$ 0.03 & 15.39 $\pm$ 0.09 \\
 & Adapt. & 18.72 $\pm$ 0.17 & 25.50 $\pm$ 0.53 & 29.49 $\pm$ 0.14 & 1.18 $\pm$ 0.09 & 25.81 $\pm$ 0.25 & 9.84 $\pm$ 0.01 \\
\bottomrule
\end{tabular}

\caption{
Main results under TRACE.
All metrics are reported on a 0-100 scale using the official metric of each benchmark.
Overall is the mean of the General, Adapt.\ and Challenge scores.
}
\label{tab:main_results}
\vspace{-5mm}
\end{table*}

\subsection{Why Group Tasks This Way?}
The goal of separating the task into three categories is simple: \emph{we want to categorize the model's knowledge and adaptation}.
When the model performs higher than $\tau$, although it was trained only on a generic dataset, that means those datasets are easy to solve and do not require any specific knowledge from models or datasets.
In contrast, when models perform poorly, it emphasizes the challenge of those tasks and datasets that models might require high robustness to achieve a high score.
On the other hand, when we use the provided training data, it tests the adaptation of models, instead of generalizing or robustness. 
In real-world scenarios, when the target task is easy, we can use the general group's performance as the indicator, while we can use the challenge group's performance to measure the robustness.
In addition, when the training data of the target task is available, we can use the adaptation group's performance as the indicator, in this case.
This makes TRACE suitable for real-world scenarios when applying text-vision embedding models that require interpretability for understanding the model improvement and flaws.

\section{Experimental Setup and Results}
\label{sec:results}

\subsection{Model Training and Evaluation}

For efficiency, we train CLIP-KD~\cite{clipkd} and TinyCLIP~\cite{tinyclip}, two famous efficiency CLIP-style models with fewer than 40M parameters, as the main models.
For each model, we keep the original architecture and training objective.
We trained three versions of those models as follows:
(i) \textbf{General} models are trained on CC12M~\cite{cc12m}.
(ii) \textbf{Adaptability} models are trained only on the 5.5M samples.
(ii) \textbf{General+Adaptability} models are combined with both General and Adaptability data.
%
%
We trained each version for three seeds.
Unless otherwise stated, we follow the original hyperparameters of each method.
Detailed training settings are provided in Appendix~\ref{sec:experiments_appendix}.




For the evaluation, we evaluate all models using the official MMEB and MIEB evaluation code.
For each task, we follow the original benchmark setting and report the official metric.
All metrics are reported on a 0-100 scale.
We evaluate results separately for the General, Adaptability, and Challenge groups as mentioned in Section~\ref{subsec:grouping}.


\subsection{Results: Category Performance}

Table~\ref{tab:main_results} reports the main results under TRACE.
%
%
TRACE produces notably different overall scores from MIEB and MMEB by explicitly separating General, Adaptation, and Challenge, with equal weight in the overall score. 
Unlike conventional aggregate scores, this design highlights the difficulty of Challenge tasks, where even teacher models achieve only 16.08–18.74, while students score below 5.0. 
Furthermore, separating General and Adaptation helps identify whether improvements arise from generalizable modeling capabilities or adaptation data. 
Adding adaptation data improves Adaptation by 7.57 and 7.01 points for CLIP-KD and TinyCLIP, respectively, while General performance remains nearly unchanged. This suggests that the improvements primarily reflect data adaptation rather than broader generalization.

\emph{Can additional training data improve model performance?} Our results suggest that improvements are possible, but not guaranteed. 
Training solely on General data preserves general capabilities but provides limited adaptability, whereas Adaptation-only training substantially degrades general performance. 
This mirrors prior findings that CLIP's generalization depends primarily on the diversity of its training data~\cite{seacliptiny}, which specialized Adaption data alone cannot provide.
Combining both (Gen.+Adapt.) achieves the best Overall performance, improving Adaptation while largely preserving General capabilities. 
However, Challenge performance remains consistently low (below 5.0), regardless of the training data. 
This suggests that data adaptation alone is insufficient to address challenging tasks, motivating research into new methods beyond data-centric improvements.


\subsection{Results: Task Performance}

We further examine task-specific difficulties and the impact of adaptation data in Table~\ref{tab:detail_benchmark}. 
In General, STS remains the most challenging task (18.2-20.1), while in Adaptation, VQA/Choice consistently shows the lowest performance (1.7-3.4). 
Adding adaptation data substantially improves Adaptation, particularly Classification (+13.8/+13.4), but provides limited gains on General (+0.1/+1.0). 
In contrast, all Challenge task types remain consistently difficult, with negligible average improvements (+0.3/+0.5). 
These findings suggest that while additional data can address certain task-specific weaknesses, neither general-purpose nor adaptation-specific training data effectively resolves the difficulties in Challenge, highlighting the need for approaches beyond data adaptation.

\begin{table}[h!]
\centering
\setlength{\tabcolsep}{2pt}
\scalebox{0.7}{
\begin{tabular}{l l c c c c c}
\toprule
& & & \multicolumn{2}{c}{\textbf{CLIP-KD}} & \multicolumn{2}{c}{\textbf{TinyCLIP}} \\
\cmidrule(lr){4-5} \cmidrule(lr){6-7}
\textbf{Cat.} & \textbf{Task Type} & \textbf{\#Tasks} & Gen. & Gen.+Adapt. & Gen. & Gen.+Adapt. \\
\midrule
\multirow{6}{*}{Gen.} & Retrieval & 23 & 36.3 & 36.9 {\footnotesize(+0.6)} & 32.2 & 33.3 {\footnotesize(+1.1)} \\
 & Classification & 3 & 32.8 & 33.9 {\footnotesize(+1.1)} & 26.9 & 31.4 {\footnotesize(+4.5)} \\
 & VQA / Choice & 14 & 42.5 & 42.7 {\footnotesize(+0.2)} & 39.8 & 40.1 {\footnotesize(+0.3)} \\
 & Grounding & 0 & -- & -- & -- & -- \\
 & STS & 8 & 20.1 & 18.2 {\footnotesize(-1.9)} & 18.5 & 19.2 {\footnotesize(+0.7)} \\
 & \bf{Average} & 48 & 35.2 & 35.3 {\footnotesize(+0.1)} & 31.8 & 32.8 {\footnotesize(+1.0)} \\
\midrule
\multirow{6}{*}{Adapt.} & Retrieval & 29 & 31.9 & 34.3 {\footnotesize(+2.4)} & 26.4 & 28.3 {\footnotesize(+1.9)} \\
 & Classification & 33 & 42.5 & 56.3 {\footnotesize(+13.8)} & 36.8 & 50.1 {\footnotesize(+13.4)} \\
 & VQA / Choice & 6 & 3.2 & 3.4 {\footnotesize(+0.2)} & 1.7 & 1.7 {\footnotesize(+0.0)} \\
 & Grounding & 4 & 34.5 & 39.2 {\footnotesize(+4.7)} & 39.3 & 41.5 {\footnotesize(+2.2)} \\
 & STS & 0 & -- & -- & -- & -- \\
 & \bf{Average} & 72 & 34.5 & 42.1 {\footnotesize(+7.6)} & 29.8 & 36.8 {\footnotesize(+7.0)} \\
\midrule
\multirow{6}{*}{Cha.} & Retrieval & 45 & 2.3 & 2.6 {\footnotesize(+0.3)} & 1.5 & 2.0 {\footnotesize(+0.5)} \\
 & Classification & 2 & 2.8 & 2.3 {\footnotesize(-0.5)} & 1.6 & 1.9 {\footnotesize(+0.3)} \\
 & VQA / Choice & 3 & 2.5 & 2.8 {\footnotesize(+0.2)} & 2.4 & 2.0 {\footnotesize(-0.4)} \\
 & Grounding & 0 & -- & -- & -- & -- \\
 & STS & 1 & 6.1 & 4.0 {\footnotesize(-2.1)} & -0.4 & 0.7 {\footnotesize(+1.1)} \\
 & \bf{Average} & 51 & 2.4 & 2.7 {\footnotesize(+0.3)} & 1.6 & 2.0 {\footnotesize(+0.5)} \\
\bottomrule
\end{tabular}}
\vspace{-2mm}
\caption{
Breakdown of the results by task type (3-seed means, 0-100).
The value in parentheses is the gain of Gen.+Adapt.\ over General.
}
\label{tab:detail_benchmark}
\vspace{-5mm}
\end{table}

\subsection{Why Is the Challenge Benchmark Difficult?}

Two patterns emerge among the 51 Challenge tasks. 
First, four of the six multilingual tasks fall into Challenge, including all three multilingual document sets and XM3600, suggesting limited multilingual generalization under predominantly English training. 
%
Second, 21 of 27 MMEB VisDoc tasks belong to Challenge, revealing persistent weaknesses in document understanding despite adaptation.

Beyond data limitations, the teacher models themselves perform poorly on Challenge (16.08–18.74), providing weak supervision for the CLIP-KD and TinyCLIP students. 
Consequently, adding adaptation data yields only marginal Challenge improvements (+0.3/+0.5), compared to substantial gains on Adaptation (+7.6/+7.0). 
These findings emphasize that existing data-centric strategies and distillation methods are insufficient for challenging out-of-distribution tasks.
%



\section{Conclusion}
\label{sec:conclusion}


%

In this work, we introduce \textbf{TRACE}, a training-aware evaluation framework for text-vision embedding models. 
Unlike existing benchmarks that primarily report aggregate performance, TRACE reorganizes 168 tasks from MMEB and MIEB into \emph{General}, \emph{Adaptability}, and \emph{Challenge} categories. 
This enables a more interpretable evaluation of model capabilities, distinguishing generalization, data adaptation, and unresolved challenges.
Our experiments demonstrate that combining general and adaptation data improves Adaptability while preserving General capabilities. 
However, Challenge tasks remain difficult even after adaptation, particularly multilingual and visual-document tasks. 
These findings reveal limitations of current small CLIP models and suggest that data-centric improvements alone are insufficient. 
TRACE provides a diagnostic evaluation framework and highlights the need for new methods to improve robustness beyond data adaptation.

\section*{Limitations}
Our study focuses on two compact CLIP models, CLIP-KD and TinyCLIP, which may limit generalization to other architectures and larger models. 
Moreover, TRACE's task categorization depends on the performance of selected reference models, meaning the General and Challenge groups may change as stronger models emerge. 
Finally, our experiments primarily use English training data, with limited data in other languages, constraining conclusions about multilingual capabilities.
Future work should extend TRACE to more model families, multilingual training settings, and diverse evaluation tasks to further validate its generalizability.

\section*{Acknowledgments}
This project is supported by the National Research Foundation, Singapore under its National Large Language Models Funding Initiative. Any opinions, findings and conclusions or recommendations expressed in this material are those of the author(s) and do not reflect the views of National Research Foundation, Singapore.




\bibliography{custom}

\appendix




\section{Related Work}
\label{sec:related-work}

\subsection{Text-Vision Embedding Models}

CLIP-style models learn transferable vision-language representations by aligning images and texts in a shared embedding space~\cite{clip}. However, their computational cost has motivated research on efficient training and compact architectures. EfficientCLIP~\cite{efficientclip}, RECLIP~\cite{reclip}, and FastCLIP~\cite{fastclip} reduce training costs, while LAION~\cite{laionb} and DataComp~\cite{datacomp} highlight the importance of training data quality. Another direction focuses on knowledge distillation, including CLIP-KD~\cite{clipkd}, TinyCLIP~\cite{tinyclip}, MobileCLIP~\cite{mobileclip}, and ComKD-CLIP~\cite{comkdclip}, which improve the efficiency-performance trade-off of compact models.
%
Our work used CLIP-KD and TinyCLIP as instruments to show how TRACE reveal where models generalize, benefit from adaptation, or remain challenged.

\subsection{Text-Vision Embedding Benchmarks}

Early CLIP evaluations primarily focused on retrieval and zero-shot classification, using benchmarks such as MS-COCO~\cite{mscoco}, Flickr30K~\cite{flickr30k}, and ImageNet~\cite{imagenet}. Recent benchmarks, including MMEB~\cite{mmeb}, MMEB-V2~\cite{mmeb_v2}, MMEB-V3~\cite{mmeb_v3}, and MIEB~\cite{mieb}, expand evaluation to diverse vision-language embedding tasks.
However, these benchmarks do not explicitly distinguish generalization from data-specific adaptation, potentially obscuring the source of performance gains. To address this, we reorganize existing benchmark datasets into \textbf{General}, \textbf{Adaptability}, and \textbf{Challenge} based on their relationship with training data. This training-aware evaluation reveals where compact CLIP models generalize, benefit from adaptation, or continue to struggle.

\section{Choice and Analysis of $\tau$}
\label{sec:tau}

\begin{figure}[h!]
  \centering
  \includegraphics[width=\linewidth]{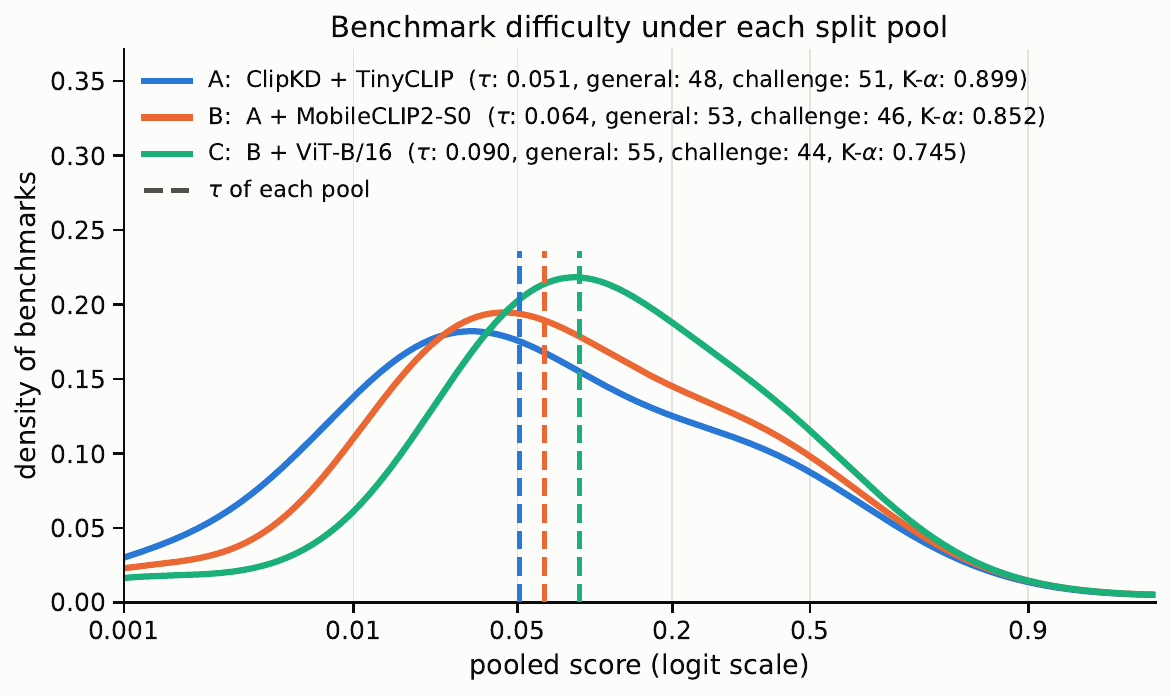}
  \caption{Kernel density estimate of the task score (logit scale) over the
  99 tasks without training data, under three pools of controlled models.
  Dashed lines mark each pool's threshold $\tau$, the median of the mixture of
  per-task densities with the pool's models as annotators. The legend gives
  $\tau$, the resulting General and Challenge counts, and Krippendorff's
  $\alpha$ among the pool's models, each splitting the tasks at its own median
  (over the 99 tasks scored by every model of the pool).
  Pool~A (CLIP-KD and TinyCLIP) is the split used in TRACE.}
  \label{fig:split_agreement_kde}
\end{figure}

To study the sensitivity of the split to the models that define $\tau$, we compare three pools of controlled models.
\textbf{A}: CLIP-KD + TinyCLIP (the TRACE split): the released CLIP-KD ViT-T/16~\cite{clipkd} and TinyCLIP ViT-8M/16 (YFCC15M)~\cite{tinyclip} checkpoints together with the three-seed mean of our General students of each family.
\textbf{B}: A + MobileCLIP2-S0~\cite{mobileclip2}.
\textbf{C}: B + CLIP ViT-B/16 (LAION-2B)~\cite{radford2021learning}, the TinyCLIP teacher.
The released weights are taken from the authors' public repositories.\footnote{
    \href{https://github.com/winycg/CLIP-KD/releases/download/CLIP-KDv0.1/ViT_T_16_cc3m_12m_ep32.pt}{CLIP-KD};
    \href{https://huggingface.co/wkcn/TinyCLIP-ViT-8M-16-Text-3M-YFCC15M}{TinyCLIP};
    \href{https://huggingface.co/apple/MobileCLIP2-S0}{MobileCLIP2-S0};
    \href{https://huggingface.co/laion/CLIP-ViT-B-16-laion2B-s34B-b88K}{CLIP ViT-B/16}.
}
As shown in Figure~\ref{fig:split_agreement_kde}, $\tau$ values are 0.050, 0.065, and 0.091 for pools A, B, and C, respectively.

\noindent
\textit{Is $\tau$ sensitive?}
To answer this, as shown in Table~\ref{tab:split_robustness}, although the number of tasks in the General and Challenge benchmarks changes since the $\tau$ is changed, the performance of General models, Adaptability models, and General+Adaptability models does not change much. 
In particular, when we compare the A, B, and C on the challenge set, there is no significant change at all.
All results are still less than 3 points.
This emphasizes the robustness of the method for choosing  $\tau$ for dataset selection. 
The detailed performance of pool A (TRACE) for each task for each controlled model for General, Adaptability, and Challenge benchmarks is in Table~\ref{tab:group_scores_general},~\ref{tab:group_scores_adaptability}, and~\ref{tab:group_scores_challenge}, respectively. 
The detailed performance for each task for A, B, and C is in Table~\ref{tab:group_members},~\ref{tab:group_members_B}, and~\ref{tab:group_members_C}, respectively.

\begin{table}[h!]
\centering
\setlength{\tabcolsep}{2pt}
\scalebox{0.7}{
\begin{tabular}{l l | c c c | c | c c c}
\toprule
& & \multicolumn{3}{c|}{\textbf{General}} & \textbf{Adapt.} & \multicolumn{3}{c}{\textbf{Challenge}} \\
\textbf{Model} & \textbf{Training} & A & B & C & & A & B & C \\
\midrule
$\tau$ & & 0.051 & 0.064 & 0.090 & -- & 0.051 & 0.064 & 0.090 \\
\#tasks & & 48 & 53 & 55 & 72 & 51 & 46 & 44 \\
\midrule
\multirow{3}{*}{ClipKD} & General & 35.2 & 32.2 & 31.1 & 34.5 & 2.4 & 2.2 & 2.3 \\
 & Gen.+Adapt. & 35.3 & 32.4 & 31.4 & 42.1 & 2.7 & 2.4 & 2.4 \\
 & Adapt. & 19.5 & 17.8 & 17.1 & 14.5 & 0.7 & 0.7 & 0.8 \\
\hline
\multirow{3}{*}{TinyClip} & General & 31.8 & 28.9 & 27.9 & 29.8 & 1.6 & 1.6 & 1.7 \\
 & Gen.+Adapt. & 32.8 & 30.0 & 29.0 & 36.8 & 2.0 & 1.9 & 1.9 \\
 & Adapt. & 25.5 & 23.2 & 22.3 & 29.5 & 1.2 & 1.1 & 1.3 \\
\bottomrule

\bottomrule
\end{tabular}}
\caption{
Robustness of the TRACE group refers to the pool of models that defines $\tau$.
\textbf{A}: CLIP-KD + TinyCLIP (the TRACE split); \textbf{B}: A + MobileCLIP2-S0; \textbf{C}: B + CLIP ViT-B/16.
Scores are 3-seed means on a 0-100 scale; Adaptability is fixed by the training data and does not depend on the pool.
}
\label{tab:split_robustness}
\end{table}

\section{Training Configuration and Computational Resource}
\label{sec:experiments_appendix}
We train models with the original code with only changes in the teacher models, student models, and training data. 
For CLIP-KD, we train student model is ViT-T-16 with teacher model ViT\_B\_16-laion400m\_e32\footnote{\href{https://github.com/winycg/CLIP-KD/releases/download/CLIP-KDv0.1/ViT_B_16-laion400m_e32.pt}{ViT-B/16 pretrained checkpoint}} using original training script\footnote{\href{https://github.com/winycg/CLIP-KD/blob/main/script/baseline/ViT_T_16_baseline.sh}{CLIP-KD training script}}. 
For TinyCLIP, we train student model is TinyCLIP-ViT-8M-16-Text-3M with teacher model ViT-B-16@laion2bh\_s34b\_b88k\footnote{\href{https://huggingface.co/laion/CLIP-ViT-B-16-laion2B-s34B-b88K}{ViT-B-16@laion2bh\_s34b\_b88k pretrained checkpoint}} using original training script\footnote{\href{https://github.com/wkcn/TinyCLIP/blob/main/script/manual_weight_inherit_100to75.sh}{TinyCLIP training script stage 1}}
\footnote{\href{https://github.com/wkcn/TinyCLIP/blob/main/script/manual_weight_inherit_75to50.sh}{TinyCLIP training script stage 2}}.
The models' training and evaluation are done on an H100 GPU. For training, the GPU hours are 2,706. For evaluation, the GPU hours are 185.

\begin{table*}[p]
\centering
\setlength{\tabcolsep}{3pt}
\scriptsize
\resizebox{\linewidth}{!}{%
\begin{tabular}[t]{l l r r r r}
\toprule
\textbf{Task} & \textbf{Suite} & \textbf{CLIP-KD} & \textbf{TinyCLIP} & \textbf{MobileCLIP2-S0} & \textbf{ViT-B/16} \\
\midrule
RP2kI2IRetrieval & MIEB & 100.0 & 100.0 & 100.0 & 100.0 \\
SOPI2IRetrieval & MIEB & 100.0 & 100.0 & 100.0 & 100.0 \\
MemotionT2IRetrieval & MIEB & 95.8 & 99.6 & 99.6 & 99.7 \\
MemotionI2TRetrieval & MIEB & 64.6 & 97.2 & 99.6 & 99.8 \\
SugarCrepe & MIEB & 76.6 & 74.7 & 82.6 & 83.5 \\
BLINKIT2IMultiChoice & MIEB & 72.6 & 74.4 & 75.1 & 73.1 \\
Flickr30kT2IRetrieval & MIEB & 67.8 & 60.8 & 81.5 & 81.3 \\
TUBerlinT2IRetrieval & MIEB & 73.1 & 29.2 & 94.1 & 91.0 \\
Flickr30kI2TRetrieval & MIEB & 61.0 & 57.3 & 79.8 & 80.1 \\
AROVisualAttribution & MIEB & 56.4 & 55.6 & 58.5 & 59.4 \\
CVBenchDepth & MIEB & 47.2 & 52.7 & 48.2 & 51.8 \\
CVBenchDistance & MIEB & 45.2 & 50.8 & 48.0 & 51.2 \\
AROVisualRelation & MIEB & 47.8 & 52.0 & 48.8 & 47.1 \\
CVBenchRelation & MIEB & 49.2 & 46.6 & 49.5 & 50.5 \\
GQA & MMEB-img & 46.7 & 45.2 & 43.9 & 52.4 \\
BLINKIT2TMultiChoice & MIEB & 41.1 & 38.2 & 39.1 & 38.5 \\
STS17MultilingualVisualSTS & MIEB & 37.2 & 38.4 & 61.3 & 79.2 \\
VisualSTS17Eng & MIEB & 37.2 & 38.4 & 61.3 & 79.2 \\
ImageNet-R & MMEB-img & 41.3 & 24.7 & 76.1 & 83.2 \\
EDIS & MMEB-img & 35.1 & 36.1 & 27.3 & 77.0 \\
Place365 & MMEB-img & 36.0 & 34.5 & 41.6 & 37.8 \\
AROFlickrOrder & MIEB & 41.8 & 28.0 & 38.4 & 36.0 \\
BLINKIT2IRetrieval & MIEB & 28.6 & 30.9 & 35.8 & 34.6 \\
AROCocoOrder & MIEB & 37.3 & 24.1 & 30.9 & 29.8 \\
\bottomrule
\end{tabular}
\hspace{1.5em}
\begin{tabular}[t]{l l r r r r}
\toprule
\textbf{Task} & \textbf{Suite} & \textbf{CLIP-KD} & \textbf{TinyCLIP} & \textbf{MobileCLIP2-S0} & \textbf{ViT-B/16} \\
\midrule
GLDv2I2TRetrieval & MIEB & 25.3 & 44.8 & 65.0 & 63.7 \\
CVBenchCount & MIEB & 61.6 & 7.6 & 37.1 & 21.3 \\
STS16VisualSTS & MIEB & 23.0 & 20.4 & 47.4 & 67.1 \\
STS12VisualSTS & MIEB & 15.1 & 24.4 & 45.5 & 63.1 \\
ObjectNet & MMEB-img & 21.5 & 17.2 & 55.2 & 51.2 \\
ROxfordEasyI2IRetrieval & MIEB & 18.3 & 23.0 & 27.8 & 28.9 \\
SketchyI2IRetrieval & MIEB & 23.7 & 3.7 & 70.9 & 57.3 \\
FORBI2IRetrieval & MIEB & 15.6 & 11.2 & 34.5 & 43.2 \\
STS15VisualSTS & MIEB & 16.9 & 12.2 & 50.1 & 74.8 \\
ImageCoDe & MIEB & 12.8 & 12.4 & 12.1 & 13.2 \\
BLINKIT2TRetrieval & MIEB & 17.6 & 9.7 & 19.7 & 15.9 \\
FashionIQ & MMEB-img & 15.2 & 5.8 & 17.5 & 13.6 \\
EDIST2ITRetrieval & MIEB & 10.9 & 10.4 & 12.2 & 21.6 \\
WITT2IRetrieval & MIEB & 9.9 & 11.8 & -- & 27.1 \\
GLDv2I2IRetrieval & MIEB & 8.3 & 13.6 & 18.6 & 21.4 \\
ROxfordMediumI2IRetrieval & MIEB & 8.4 & 11.0 & 15.9 & 15.5 \\
STS13VisualSTS & MIEB & 10.8 & 4.8 & 35.7 & 61.2 \\
RParisEasyI2IRetrieval$^{C}$ & MIEB & 7.5 & 8.0 & 8.5 & 8.2 \\
ImageCoDeT2IRetrieval$^{C}$ & MIEB & 7.6 & 6.5 & 10.0 & 10.4 \\
OVEN$^{B}$ & MMEB-img & 6.3 & 7.3 & 2.0 & 45.0 \\
STSBenchmarkMultilingualVisualSTS & MIEB & 5.3 & 5.9 & 28.9 & 57.6 \\
VisualSTS-b-Eng & MIEB & 5.3 & 5.9 & 28.9 & 57.6 \\
Winoground$^{C}$ & MIEB & 7.5 & 5.2 & 9.0 & 8.2 \\
ViDoSeek-doc & MMEB-doc & 6.7 & 4.0 & 37.2 & 68.2 \\
\bottomrule
\end{tabular}
}
\caption{
Scores (0--100) of the original models on every task of the TRACE General group (48 tasks)
under the CLIP-KD + TinyCLIP split (pool A), sorted by the pooled score of the pool's models. A superscript marks tasks that move to the other non-adaptable group when $\tau$ is recomputed with pool B (+ MobileCLIP2-S0) or pool C (+ MobileCLIP2-S0 + ViT-B/16).
}
\label{tab:group_scores_general}
\end{table*}

\begin{table*}[p]
\centering
\setlength{\tabcolsep}{3pt}
\scriptsize
\resizebox{\linewidth}{!}{%
\begin{tabular}[t]{l l r r r r}
\toprule
\textbf{Task} & \textbf{Suite} & \textbf{CLIP-KD} & \textbf{TinyCLIP} & \textbf{MobileCLIP2-S0} & \textbf{ViT-B/16} \\
\midrule
SciMMIRI2TRetrieval & MIEB & 100.0 & 100.0 & 100.0 & 100.0 \\
SciMMIRT2IRetrieval & MIEB & 100.0 & 100.0 & 100.0 & 100.0 \\
STL10ZeroShot & MIEB & 90.9 & 92.4 & 96.4 & 97.8 \\
ImageNet10Clustering & MIEB & 88.3 & 82.0 & 92.7 & 96.9 \\
HatefulMemesI2TRetrieval & MIEB & 82.9 & 82.9 & 82.9 & 82.9 \\
HatefulMemesT2IRetrieval & MIEB & 82.9 & 82.9 & 82.9 & 82.9 \\
CIFAR10ZeroShot & MIEB & 75.2 & 72.5 & 94.0 & 95.1 \\
Caltech101ZeroShot & MIEB & 74.8 & 70.2 & 84.5 & 81.8 \\
VOC2007 & MMEB-img & 62.2 & 65.0 & 63.9 & 52.1 \\
OxfordPetsZeroShot & MIEB & 72.4 & 46.4 & 91.7 & 90.3 \\
NIGHTS & MMEB-img & 60.3 & 56.4 & 65.9 & 66.1 \\
WebQA & MMEB-img & 64.4 & 52.8 & 56.9 & 73.1 \\
SUN397ZeroShot & MIEB & 56.8 & 55.6 & 68.5 & 70.7 \\
PatchCamelyonZeroShot & MIEB & 55.6 & 54.0 & 55.6 & 55.8 \\
SUN397 & MMEB-img & 56.3 & 54.7 & 65.6 & 68.9 \\
CIFAR10Clustering & MIEB & 50.3 & 51.1 & 87.1 & 79.7 \\
HatefulMemes & MMEB-img & 50.7 & 51.0 & 52.2 & 51.8 \\
RenderedSST2 & MIEB & 50.1 & 50.7 & 51.8 & 59.9 \\
RefCOCO-Matching & MMEB-img & 40.1 & 60.1 & 23.6 & 68.1 \\
CIFAR100Clustering & MIEB & 51.0 & 50.8 & 74.2 & 73.7 \\
ImageNetDog15Clustering & MIEB & 56.2 & 37.9 & 74.9 & 80.0 \\
Food101ZeroShot & MIEB & 47.9 & 57.0 & 84.3 & 85.6 \\
TinyImageNetClustering & MIEB & 48.1 & 47.2 & 69.6 & 67.2 \\
MSCOCO\_t2i & MMEB-img & 49.9 & 44.3 & 67.5 & 63.7 \\
MSCOCO\_i2t & MMEB-img & 49.4 & 43.0 & 65.6 & 61.8 \\
MSCOCOT2IRetrieval & MIEB & 44.1 & 37.9 & 60.3 & 58.5 \\
Imagenet1kZeroShot & MIEB & 41.1 & 40.8 & 71.7 & 68.4 \\
RefCOCO & MMEB-img & 37.9 & 40.4 & 25.0 & 54.5 \\
CIFAR100ZeroShot & MIEB & 41.2 & 39.4 & 76.7 & 76.1 \\
N24News & MMEB-img & 38.8 & 36.9 & 23.1 & 38.9 \\
ImageNet-1K & MMEB-img & 41.0 & 36.7 & 68.1 & 63.3 \\
Visual7W-Pointing & MMEB-img & 34.7 & 36.3 & 34.2 & 56.1 \\
VisualNews\_i2t & MMEB-img & 40.5 & 36.0 & 60.6 & 77.8 \\
MSCOCOI2TRetrieval & MIEB & 37.2 & 32.9 & 55.6 & 52.3 \\
VisualNews\_t2i & MMEB-img & 38.0 & 34.3 & 55.7 & 74.0 \\
RESISC45ZeroShot & MIEB & 34.2 & 28.3 & 57.8 & 65.9 \\
\bottomrule
\end{tabular}
\hspace{1.5em}
\begin{tabular}[t]{l l r r r r}
\toprule
\textbf{Task} & \textbf{Suite} & \textbf{CLIP-KD} & \textbf{TinyCLIP} & \textbf{MobileCLIP2-S0} & \textbf{ViT-B/16} \\
\midrule
EuroSATZeroShot & MIEB & 31.9 & 25.8 & 49.9 & 49.1 \\
WebQAT2ITRetrieval & MIEB & 32.9 & 25.2 & 27.2 & 40.2 \\
UCF101ZeroShot & MIEB & 27.8 & 28.2 & 53.4 & 55.1 \\
DTDZeroShot & MIEB & 29.1 & 27.0 & 59.3 & 54.0 \\
MSCOCO & MMEB-img & 26.6 & 27.9 & 13.0 & 34.7 \\
FER2013ZeroShot & MIEB & 25.0 & 22.8 & 39.4 & 49.9 \\
NIGHTSI2IRetrieval & MIEB & 23.4 & 21.2 & 26.5 & 25.6 \\
CIRR & MMEB-img & 24.8 & 14.1 & 44.7 & 15.1 \\
CLEVRZeroShot & MIEB & 15.4 & 16.0 & 17.6 & 21.1 \\
WebQAT2TRetrieval & MIEB & 24.0 & 13.2 & 19.7 & 30.8 \\
BirdsnapZeroShot & MIEB & 11.8 & 34.3 & 56.0 & 58.4 \\
CLEVRCountZeroShot & MIEB & 14.7 & 12.7 & 19.0 & 21.4 \\
VisDial & MMEB-img & 14.7 & 21.0 & 26.5 & 25.1 \\
StanfordCarsZeroShot & MIEB & 19.8 & 8.0 & 90.0 & 88.2 \\
MNISTZeroShot & MIEB & 19.9 & 11.3 & 84.7 & 68.2 \\
CIRRIT2IRetrieval & MIEB & 18.6 & 10.4 & 14.6 & 14.1 \\
XFlickr30kCoT2IRetrieval & MIEB & 10.3 & 12.0 & -- & 22.4 \\
GTSRBZeroShot & MIEB & 12.6 & 6.7 & 50.9 & 48.9 \\
Country211 & MMEB-img & 7.5 & 11.6 & 11.4 & 16.3 \\
Country211ZeroShot & MIEB & 7.9 & 10.9 & 11.0 & 18.7 \\
ViDoRe\_infovqa & MMEB-doc & 8.4 & 5.3 & 20.0 & 48.9 \\
VidoreInfoVQARetrieval & MIEB & 8.0 & 5.3 & 20.0 & 49.0 \\
OK-VQA & MMEB-img & 8.0 & 7.0 & 6.7 & 11.7 \\
VisRAG\_InfoVQA & MMEB-doc & 8.8 & 4.0 & 21.0 & 51.8 \\
VisualNewsI2TRetrieval & MIEB & 6.1 & 4.7 & 16.5 & 30.4 \\
VisualNewsT2IRetrieval & MIEB & 5.5 & 4.8 & 14.1 & 29.4 \\
InfographicsVQA & MMEB-img & 4.2 & 2.9 & 4.0 & 4.7 \\
ViDoRe\_docvqa & MMEB-doc & 3.7 & 1.9 & 8.5 & 13.6 \\
VidoreDocVQARetrieval & MIEB & 3.6 & 1.9 & 8.3 & 14.1 \\
DocVQA & MMEB-img & 2.8 & 1.4 & 4.2 & 5.5 \\
VisRAG\_MP-DocVQA & MMEB-doc & 1.9 & 1.8 & 6.7 & 19.6 \\
A-OKVQA & MMEB-img & 1.3 & 1.4 & 1.5 & 3.3 \\
Visual7W & MMEB-img & 1.2 & 1.2 & 1.1 & 2.6 \\
ChartQA & MMEB-img & 1.4 & 0.9 & 0.9 & 1.6 \\
VisRAG\_ChartQA & MMEB-doc & 0.6 & 0.0 & 7.8 & 42.7 \\
StanfordCarsI2IRetrieval & MIEB & 0.0 & 0.0 & 0.0 & 0.0 \\
\bottomrule
\end{tabular}
}
\caption{
Scores (0--100) of the original models on every task of the TRACE Adaptability group (72 tasks)
under the CLIP-KD + TinyCLIP split (pool A), sorted by the pooled score of the pool's models. A superscript marks tasks that move to the other non-adaptable group when $\tau$ is recomputed with pool B (+ MobileCLIP2-S0) or pool C (+ MobileCLIP2-S0 + ViT-B/16).
}
\label{tab:group_scores_adaptability}
\end{table*}

\begin{table*}[p]
\centering
\setlength{\tabcolsep}{3pt}
\scriptsize
\resizebox{\linewidth}{!}{%
\begin{tabular}[t]{l l r r r r}
\toprule
\textbf{Task} & \textbf{Suite} & \textbf{CLIP-KD} & \textbf{TinyCLIP} & \textbf{MobileCLIP2-S0} & \textbf{ViT-B/16} \\
\midrule
ViDoRe\_arxivqa$^{BC}$ & MMEB-doc & 6.5 & 3.7 & 16.4 & 28.1 \\
VidoreArxivQARetrieval$^{BC}$ & MIEB & 6.4 & 3.8 & 16.4 & 28.1 \\
VidoreTabfquadRetrieval & MIEB & 3.7 & 4.7 & 11.7 & 23.1 \\
ViDoRe\_tabfquad & MMEB-doc & 3.5 & 4.7 & 11.7 & 23.8 \\
MMLongBench-doc$^{C}$ & MMEB-doc & 4.8 & 3.6 & 11.8 & 26.4 \\
FGVCAircraftZeroShot$^{BC}$ & MIEB & 3.1 & 7.0 & 21.0 & 26.8 \\
XM3600T2IRetrieval & MIEB & 3.8 & 5.5 & -- & 11.3 \\
Wiki-SS-NQ$^{C}$ & MMEB-img & 5.0 & 2.7 & 10.5 & 44.6 \\
ScienceQA & MMEB-img & 3.6 & 3.5 & 5.4 & 10.0 \\
ViDoRe\_economics\_reports\_v2 & MMEB-doc & 3.1 & 3.5 & 3.2 & 19.9 \\
ViDoRe\_syntheticDocQA\_artificial\_intelligence & MMEB-doc & 5.1 & 2.1 & 7.2 & 26.4 \\
VidoreSyntheticDocQAAIRetrieval & MIEB & 5.1 & 2.1 & 7.2 & 26.4 \\
VisRAG\_SlideVQA$^{BC}$ & MMEB-doc & 4.3 & 1.7 & 30.9 & 62.3 \\
ViDoRe\_economics\_reports\_v2\_multilingual & MMEB-doc & 2.5 & 2.6 & 3.2 & 9.7 \\
ViDoRe\_esg\_reports\_v2 & MMEB-doc & 4.0 & 1.1 & 6.1 & 12.9 \\
TextVQA & MMEB-img & 2.2 & 3.2 & 1.7 & 10.8 \\
RParisMediumI2IRetrieval & MIEB & 2.5 & 2.5 & 2.6 & 2.6 \\
FashionIQIT2IRetrieval & MIEB & 4.4 & 1.0 & 4.2 & 4.8 \\
ImageNet-A & MMEB-img & 1.6 & 4.4 & 6.1 & 14.4 \\
VizWizIT2TRetrieval & MIEB & 2.0 & 2.7 & 4.8 & 4.8 \\
ViDoRe\_syntheticDocQA\_energy$^{C}$ & MMEB-doc & 2.0 & 1.0 & 7.9 & 39.5 \\
InfoSeekIT2ITRetrieval & MIEB & 0.5 & 3.4 & 3.8 & 8.6 \\
VidoreSyntheticDocQAEnergyRetrieval$^{C}$ & MIEB & 2.0 & 1.0 & 7.9 & 39.4 \\
OVENIT2ITRetrieval & MIEB & 1.5 & 1.9 & 4.9 & 7.7 \\
STS14VisualSTS$^{BC}$ & MIEB & 2.2 & 0.4 & 37.7 & 60.3 \\
ViDoSeek-page$^{BC}$ & MMEB-doc & 2.4 & 1.3 & 34.1 & 60.3 \\
\bottomrule
\end{tabular}
\hspace{1.5em}
\begin{tabular}[t]{l l r r r r}
\toprule
\textbf{Task} & \textbf{Suite} & \textbf{CLIP-KD} & \textbf{TinyCLIP} & \textbf{MobileCLIP2-S0} & \textbf{ViT-B/16} \\
\midrule
MMLongBench-page & MMEB-doc & 2.5 & 1.0 & 10.4 & 20.7 \\
ROxfordHardI2IRetrieval & MIEB & 1.1 & 1.1 & 3.8 & 3.4 \\
VidoreTatdqaRetrieval & MIEB & 1.8 & 1.1 & 2.5 & 10.1 \\
ViDoRe\_tatdqa & MMEB-doc & 1.7 & 1.1 & 2.5 & 9.6 \\
ViDoRe\_esg\_reports\_human\_labeled\_v2 & MMEB-doc & 2.1 & 1.9 & 3.1 & 11.7 \\
Fashion200kI2TRetrieval & MIEB & 1.7 & 0.2 & 8.3 & 10.3 \\
ViDoRe\_shiftproject & MMEB-doc & 2.1 & 1.4 & 2.5 & 8.8 \\
VidoreShiftProjectRetrieval & MIEB & 2.1 & 1.4 & 2.5 & 9.4 \\
ViDoRe\_esg\_reports\_v2\_multilingual & MMEB-doc & 1.4 & 0.6 & 3.8 & 5.8 \\
VizWiz & MMEB-img & 0.8 & 1.4 & 1.9 & 6.3 \\
Fashion200kT2IRetrieval & MIEB & 1.3 & 0.1 & 6.9 & 7.9 \\
ViDoRe\_biomedical\_lectures\_v2 & MMEB-doc & 1.3 & 0.4 & 10.5 & 33.0 \\
VidoreSyntheticDocQAGovernmentReportsRetrieval & MIEB & 1.0 & 1.0 & 3.0 & 31.3 \\
ViDoRe\_syntheticDocQA\_government\_reports & MMEB-doc & 1.0 & 1.0 & 3.0 & 31.9 \\
VQA2IT2TRetrieval & MIEB & 0.8 & 1.0 & 0.6 & 1.5 \\
VisRAG\_ArxivQA & MMEB-doc & 1.0 & 0.5 & 8.6 & 18.9 \\
ViDoRe\_biomedical\_lectures\_v2\_multilingual & MMEB-doc & 0.6 & 0.4 & 6.1 & 16.3 \\
VidoreSyntheticDocQAHealthcareIndustryRetrieval & MIEB & 0.0 & 0.5 & 8.0 & 38.2 \\
ViDoRe\_syntheticDocQA\_healthcare\_industry & MMEB-doc & 0.0 & 0.5 & 8.0 & 38.8 \\
OVENIT2TRetrieval & MIEB & 0.3 & 0.2 & 0.7 & 2.0 \\
InfoSeekIT2TRetrieval & MIEB & 0.1 & 0.1 & 0.3 & 0.8 \\
RParisHardI2IRetrieval & MIEB & 0.1 & 0.1 & 0.1 & 0.1 \\
VisRAG\_PlotQA & MMEB-doc & 0.1 & 0.0 & 0.2 & 5.7 \\
CUB200I2IRetrieval & MIEB & 0.0 & 0.0 & 0.0 & 0.0 \\
METI2IRetrieval & MIEB & 0.0 & 0.0 & 0.0 & 0.0 \\
\bottomrule
\end{tabular}
}
\caption{
Scores (0--100) of the original models on every task of the TRACE Challenge group (51 tasks)
under the CLIP-KD + TinyCLIP split (pool A), sorted by the pooled score of the pool's models. A superscript marks tasks that move to the other non-adaptable group when $\tau$ is recomputed with pool B (+ MobileCLIP2-S0) or pool C (+ MobileCLIP2-S0 + ViT-B/16).
}
\label{tab:group_scores_challenge}
\end{table*}

\begin{table*}[p]
\centering
\setlength{\tabcolsep}{3pt}
\scriptsize
\resizebox{\linewidth}{!}{%
\begin{tabular}[t]{l l r}
\toprule
\multicolumn{3}{l}{\textbf{General} (48 tasks)} \\
\textbf{Task} & \textbf{Suite} & \textbf{Score} \\
\midrule
RP2kI2IRetrieval & MIEB & 100.0 \\
SOPI2IRetrieval & MIEB & 100.0 \\
MemotionT2IRetrieval & MIEB & 98.6 \\
MemotionI2TRetrieval & MIEB & 89.2 \\
SugarCrepe & MIEB & 75.4 \\
BLINKIT2IMultiChoice & MIEB & 73.8 \\
Flickr30kT2IRetrieval & MIEB & 62.8 \\
TUBerlinT2IRetrieval & MIEB & 57.8 \\
Flickr30kI2TRetrieval & MIEB & 56.8 \\
AROVisualAttribution & MIEB & 55.5 \\
CVBenchDepth & MIEB & 50.2 \\
CVBenchDistance & MIEB & 49.3 \\
AROVisualRelation & MIEB & 49.1 \\
CVBenchRelation & MIEB & 47.8 \\
GQA & MMEB-img & 43.5 \\
BLINKIT2TMultiChoice & MIEB & 39.8 \\
STS17MultilingualVisualSTS & MIEB & 36.9 \\
VisualSTS17Eng & MIEB & 36.9 \\
ImageNet-R & MMEB-img & 36.7 \\
EDIS & MMEB-img & 33.0 \\
Place365 & MMEB-img & 32.9 \\
AROFlickrOrder & MIEB & 32.3 \\
BLINKIT2IRetrieval & MIEB & 30.2 \\
AROCocoOrder & MIEB & 29.3 \\
GLDv2I2TRetrieval & MIEB & 27.7 \\
CVBenchCount & MIEB & 26.5 \\
STS16VisualSTS & MIEB & 21.9 \\
STS12VisualSTS & MIEB & 21.1 \\
ObjectNet & MMEB-img & 19.0 \\
ROxfordEasyI2IRetrieval & MIEB & 17.7 \\
SketchyI2IRetrieval & MIEB & 15.4 \\
FORBI2IRetrieval & MIEB & 14.5 \\
STS15VisualSTS & MIEB & 14.4 \\
ImageCoDe & MIEB & 12.6 \\
BLINKIT2TRetrieval & MIEB & 12.2 \\
FashionIQ & MMEB-img & 10.8 \\
EDIST2ITRetrieval & MIEB & 9.8 \\
WITT2IRetrieval$_{M}$ & MIEB & 9.6 \\
GLDv2I2IRetrieval & MIEB & 9.2 \\
ROxfordMediumI2IRetrieval & MIEB & 9.1 \\
STS13VisualSTS & MIEB & 8.5 \\
RParisEasyI2IRetrieval$^{C}$ & MIEB & 7.9 \\
ImageCoDeT2IRetrieval$^{C}$ & MIEB & 6.9 \\
OVEN$^{B}$ & MMEB-img & 6.6 \\
STSBenchmarkMultilingualVisualSTS & MIEB & 6.5 \\
VisualSTS-b-Eng & MIEB & 6.5 \\
Winoground$^{C}$ & MIEB & 6.0 \\
ViDoSeek-doc & MMEB-doc & 5.5 \\
\bottomrule
\end{tabular}
\hspace{1.5em}
\begin{tabular}[t]{l l r}
\toprule
\multicolumn{3}{l}{\textbf{Adaptability} (72 tasks)} \\
\textbf{Task} & \textbf{Suite} & \textbf{Score} \\
\midrule
SciMMIRI2TRetrieval & MIEB & 100.0 \\
SciMMIRT2IRetrieval & MIEB & 100.0 \\
STL10ZeroShot & MIEB & 90.6 \\
ImageNet10Clustering & MIEB & 85.4 \\
HatefulMemesI2TRetrieval & MIEB & 82.9 \\
HatefulMemesT2IRetrieval & MIEB & 82.9 \\
CIFAR10ZeroShot & MIEB & 72.6 \\
Caltech101ZeroShot & MIEB & 71.8 \\
VOC2007 & MMEB-img & 61.6 \\
OxfordPetsZeroShot & MIEB & 60.0 \\
NIGHTS & MMEB-img & 58.4 \\
WebQA & MMEB-img & 54.7 \\
SUN397ZeroShot & MIEB & 54.0 \\
PatchCamelyonZeroShot & MIEB & 53.5 \\
SUN397 & MMEB-img & 52.5 \\
CIFAR10Clustering & MIEB & 51.3 \\
HatefulMemes & MMEB-img & 50.8 \\
RenderedSST2 & MIEB & 50.4 \\
RefCOCO-Matching & MMEB-img & 50.1 \\
CIFAR100Clustering & MIEB & 49.7 \\
ImageNetDog15Clustering & MIEB & 49.3 \\
Food101ZeroShot & MIEB & 49.2 \\
TinyImageNetClustering & MIEB & 47.1 \\
MSCOCO\_t2i & MMEB-img & 45.4 \\
MSCOCO\_i2t & MMEB-img & 43.9 \\
MSCOCOT2IRetrieval & MIEB & 39.9 \\
Imagenet1kZeroShot & MIEB & 38.9 \\
RefCOCO & MMEB-img & 38.5 \\
CIFAR100ZeroShot & MIEB & 37.8 \\
N24News & MMEB-img & 36.8 \\
ImageNet-1K & MMEB-img & 35.6 \\
Visual7W-Pointing & MMEB-img & 34.6 \\
VisualNews\_i2t & MMEB-img & 34.5 \\
MSCOCOI2TRetrieval & MIEB & 33.4 \\
VisualNews\_t2i & MMEB-img & 32.7 \\
RESISC45ZeroShot & MIEB & 31.4 \\
EuroSATZeroShot & MIEB & 28.8 \\
WebQAT2ITRetrieval & MIEB & 26.6 \\
UCF101ZeroShot & MIEB & 26.6 \\
DTDZeroShot & MIEB & 26.5 \\
MSCOCO & MMEB-img & 26.4 \\
FER2013ZeroShot & MIEB & 22.8 \\
NIGHTSI2IRetrieval & MIEB & 22.5 \\
CIRR & MMEB-img & 19.8 \\
CLEVRZeroShot & MIEB & 17.6 \\
WebQAT2TRetrieval & MIEB & 17.3 \\
BirdsnapZeroShot & MIEB & 16.3 \\
CLEVRCountZeroShot & MIEB & 15.5 \\
VisDial & MMEB-img & 15.5 \\
StanfordCarsZeroShot & MIEB & 15.2 \\
MNISTZeroShot & MIEB & 13.6 \\
CIRRIT2IRetrieval & MIEB & 13.4 \\
XFlickr30kCoT2IRetrieval$_{M}$ & MIEB & 10.1 \\
GTSRBZeroShot & MIEB & 10.1 \\
Country211 & MMEB-img & 8.1 \\
Country211ZeroShot & MIEB & 8.0 \\
ViDoRe\_infovqa & MMEB-doc & 7.1 \\
VidoreInfoVQARetrieval & MIEB & 7.0 \\
OK-VQA & MMEB-img & 6.9 \\
VisRAG\_InfoVQA & MMEB-doc & 6.7 \\
VisualNewsI2TRetrieval & MIEB & 4.6 \\
VisualNewsT2IRetrieval & MIEB & 4.4 \\
InfographicsVQA & MMEB-img & 3.4 \\
ViDoRe\_docvqa & MMEB-doc & 2.6 \\
VidoreDocVQARetrieval & MIEB & 2.5 \\
DocVQA & MMEB-img & 2.4 \\
VisRAG\_MP-DocVQA & MMEB-doc & 1.9 \\
A-OKVQA & MMEB-img & 1.2 \\
Visual7W & MMEB-img & 1.1 \\
ChartQA & MMEB-img & 0.9 \\
VisRAG\_ChartQA & MMEB-doc & 0.6 \\
StanfordCarsI2IRetrieval & MIEB & 0.0 \\
\bottomrule
\end{tabular}
\hspace{1.5em}
\begin{tabular}[t]{l l r}
\toprule
\multicolumn{3}{l}{\textbf{Challenge} (51 tasks)} \\
\textbf{Task} & \textbf{Suite} & \textbf{Score} \\
\midrule
ViDoRe\_arxivqa$^{BC}$ & MMEB-doc & 4.8 \\
VidoreArxivQARetrieval$^{BC}$ & MIEB & 4.7 \\
VidoreTabfquadRetrieval & MIEB & 4.5 \\
ViDoRe\_tabfquad & MMEB-doc & 4.4 \\
MMLongBench-doc$^{C}$ & MMEB-doc & 4.3 \\
XM3600T2IRetrieval$_{M}$ & MIEB & 4.0 \\
FGVCAircraftZeroShot$^{BC}$ & MIEB & 4.0 \\
Wiki-SS-NQ$^{C}$ & MMEB-img & 3.8 \\
ScienceQA & MMEB-img & 3.8 \\
ViDoRe\_economics\_reports\_v2 & MMEB-doc & 3.6 \\
VidoreSyntheticDocQAAIRetrieval & MIEB & 3.3 \\
ViDoRe\_syntheticDocQA\_artificial\_intelligence & MMEB-doc & 3.3 \\
VisRAG\_SlideVQA$^{BC}$ & MMEB-doc & 3.1 \\
ViDoRe\_economics\_reports\_v2\_multilingual$_{M}$ & MMEB-doc & 2.7 \\
TextVQA & MMEB-img & 2.6 \\
ViDoRe\_esg\_reports\_v2 & MMEB-doc & 2.6 \\
RParisMediumI2IRetrieval & MIEB & 2.5 \\
FashionIQIT2IRetrieval & MIEB & 2.3 \\
InfoSeekIT2ITRetrieval & MIEB & 2.2 \\
VizWizIT2TRetrieval & MIEB & 2.2 \\
ImageNet-A & MMEB-img & 2.2 \\
ViDoRe\_syntheticDocQA\_energy$^{C}$ & MMEB-doc & 2.2 \\
OVENIT2ITRetrieval & MIEB & 2.1 \\
VidoreSyntheticDocQAEnergyRetrieval$^{C}$ & MIEB & 2.1 \\
STS14VisualSTS$^{BC}$ & MIEB & 2.1 \\
ViDoSeek-page$^{BC}$ & MMEB-doc & 2.0 \\
ROxfordHardI2IRetrieval & MIEB & 1.9 \\
MMLongBench-page & MMEB-doc & 1.9 \\
VidoreTatdqaRetrieval & MIEB & 1.4 \\
ViDoRe\_tatdqa & MMEB-doc & 1.4 \\
ViDoRe\_esg\_reports\_human\_labeled\_v2 & MMEB-doc & 1.3 \\
Fashion200kI2TRetrieval & MIEB & 1.2 \\
VidoreShiftProjectRetrieval & MIEB & 1.2 \\
ViDoRe\_shiftproject & MMEB-doc & 1.2 \\
ViDoRe\_esg\_reports\_v2\_multilingual$_{M}$ & MMEB-doc & 1.1 \\
VizWiz & MMEB-img & 1.0 \\
Fashion200kT2IRetrieval & MIEB & 0.9 \\
ViDoRe\_biomedical\_lectures\_v2 & MMEB-doc & 0.9 \\
VQA2IT2TRetrieval & MIEB & 0.8 \\
VidoreSyntheticDocQAGovernmentReportsRetrieval & MIEB & 0.8 \\
ViDoRe\_syntheticDocQA\_government\_reports & MMEB-doc & 0.8 \\
VisRAG\_ArxivQA & MMEB-doc & 0.8 \\
ViDoRe\_biomedical\_lectures\_v2\_multilingual$_{M}$ & MMEB-doc & 0.5 \\
VidoreSyntheticDocQAHealthcareIndustryRetrieval & MIEB & 0.4 \\
ViDoRe\_syntheticDocQA\_healthcare\_industry & MMEB-doc & 0.4 \\
OVENIT2TRetrieval & MIEB & 0.3 \\
InfoSeekIT2TRetrieval & MIEB & 0.2 \\
RParisHardI2IRetrieval & MIEB & 0.1 \\
CUB200I2IRetrieval & MIEB & 0.0 \\
METI2IRetrieval & MIEB & 0.0 \\
VisRAG\_PlotQA & MMEB-doc & 0.0 \\
\bottomrule
\end{tabular}
}
\caption{
Tasks in each TRACE group under the CLIP-KD + TinyCLIP split (pool A), sorted by the pooled score of the pool's models (Score, 0--100).
Adaptability is fixed by the availability of training data; General and Challenge are split at $\tau$. A superscript marks tasks that move to the other non-adaptable group when $\tau$ is recomputed with pool B (+ MobileCLIP2-S0, 7 tasks) or pool C (+ MobileCLIP2-S0 + ViT-B/16, 13 tasks).
A subscript $_{M}$ marks the tasks evaluated over several languages.
}
\label{tab:group_members}
\end{table*}

\begin{table*}[p]
\centering
\setlength{\tabcolsep}{3pt}
\scriptsize
\resizebox{\linewidth}{!}{%
\begin{tabular}[t]{l l r}
\toprule
\multicolumn{3}{l}{\textbf{General} (53 tasks)} \\
\textbf{Task} & \textbf{Suite} & \textbf{Score} \\
\midrule
RP2kI2IRetrieval & MIEB & 100.0 \\
SOPI2IRetrieval & MIEB & 100.0 \\
MemotionT2IRetrieval & MIEB & 98.8 \\
MemotionI2TRetrieval & MIEB & 91.3 \\
SugarCrepe & MIEB & 76.8 \\
BLINKIT2IMultiChoice & MIEB & 74.0 \\
Flickr30kT2IRetrieval & MIEB & 66.5 \\
TUBerlinT2IRetrieval & MIEB & 65.0 \\
Flickr30kI2TRetrieval & MIEB & 61.4 \\
AROVisualAttribution & MIEB & 56.1 \\
CVBenchDepth & MIEB & 49.8 \\
AROVisualRelation & MIEB & 49.0 \\
CVBenchDistance & MIEB & 49.0 \\
CVBenchRelation & MIEB & 48.2 \\
ImageNet-R & MMEB-img & 44.6 \\
GQA & MMEB-img & 43.6 \\
STS17MultilingualVisualSTS & MIEB & 41.8 \\
VisualSTS17Eng & MIEB & 41.8 \\
BLINKIT2TMultiChoice & MIEB & 39.6 \\
GLDv2I2TRetrieval & MIEB & 35.2 \\
Place365 & MMEB-img & 34.7 \\
AROFlickrOrder & MIEB & 33.5 \\
EDIS & MMEB-img & 31.8 \\
BLINKIT2IRetrieval & MIEB & 31.3 \\
AROCocoOrder & MIEB & 29.6 \\
CVBenchCount & MIEB & 28.6 \\
STS16VisualSTS & MIEB & 27.0 \\
SketchyI2IRetrieval & MIEB & 26.5 \\
ObjectNet & MMEB-img & 26.2 \\
STS12VisualSTS & MIEB & 26.0 \\
STS15VisualSTS & MIEB & 21.6 \\
ROxfordEasyI2IRetrieval & MIEB & 19.7 \\
FORBI2IRetrieval & MIEB & 18.5 \\
STS13VisualSTS & MIEB & 14.0 \\
BLINKIT2TRetrieval & MIEB & 13.7 \\
ImageCoDe & MIEB & 12.5 \\
FashionIQ & MMEB-img & 12.2 \\
ViDoSeek-doc & MMEB-doc & 11.8 \\
GLDv2I2IRetrieval & MIEB & 11.1 \\
STSBenchmarkMultilingualVisualSTS & MIEB & 10.9 \\
VisualSTS-b-Eng & MIEB & 10.9 \\
ROxfordMediumI2IRetrieval & MIEB & 10.4 \\
EDIST2ITRetrieval & MIEB & 10.3 \\
WITT2IRetrieval$_{M}$ & MIEB & 9.6 \\
STS14VisualSTS$^{A}$ & MIEB & 9.2 \\
VisRAG\_SlideVQA$^{A}$ & MMEB-doc & 8.6 \\
ViDoSeek-page$^{A}$ & MMEB-doc & 8.4 \\
RParisEasyI2IRetrieval & MIEB & 8.0 \\
ImageCoDeT2IRetrieval & MIEB & 7.5 \\
FGVCAircraftZeroShot$^{A}$ & MIEB & 7.4 \\
ViDoRe\_arxivqa$^{A}$ & MMEB-doc & 7.1 \\
VidoreArxivQARetrieval$^{A}$ & MIEB & 7.0 \\
Winoground & MIEB & 6.6 \\
\bottomrule
\end{tabular}
\hspace{1.5em}
\begin{tabular}[t]{l l r}
\toprule
\multicolumn{3}{l}{\textbf{Adaptability} (72 tasks)} \\
\textbf{Task} & \textbf{Suite} & \textbf{Score} \\
\midrule
SciMMIRI2TRetrieval & MIEB & 100.0 \\
SciMMIRT2IRetrieval & MIEB & 100.0 \\
STL10ZeroShot & MIEB & 91.7 \\
ImageNet10Clustering & MIEB & 86.9 \\
HatefulMemesI2TRetrieval & MIEB & 82.9 \\
HatefulMemesT2IRetrieval & MIEB & 82.9 \\
CIFAR10ZeroShot & MIEB & 76.9 \\
Caltech101ZeroShot & MIEB & 74.3 \\
OxfordPetsZeroShot & MIEB & 66.3 \\
VOC2007 & MMEB-img & 62.0 \\
NIGHTS & MMEB-img & 59.9 \\
CIFAR10Clustering & MIEB & 58.4 \\
SUN397ZeroShot & MIEB & 56.9 \\
Food101ZeroShot & MIEB & 56.2 \\
WebQA & MMEB-img & 55.2 \\
SUN397 & MMEB-img & 55.1 \\
CIFAR100Clustering & MIEB & 54.6 \\
ImageNetDog15Clustering & MIEB & 54.4 \\
PatchCamelyonZeroShot & MIEB & 53.9 \\
TinyImageNetClustering & MIEB & 51.6 \\
HatefulMemes & MMEB-img & 51.1 \\
RenderedSST2 & MIEB & 50.7 \\
MSCOCO\_t2i & MMEB-img & 49.9 \\
MSCOCO\_i2t & MMEB-img & 48.2 \\
CIFAR100ZeroShot & MIEB & 45.5 \\
Imagenet1kZeroShot & MIEB & 45.5 \\
RefCOCO-Matching & MMEB-img & 44.8 \\
MSCOCOT2IRetrieval & MIEB & 44.0 \\
ImageNet-1K & MMEB-img & 42.1 \\
VisualNews\_i2t & MMEB-img & 39.7 \\
MSCOCOI2TRetrieval & MIEB & 37.8 \\
VisualNews\_t2i & MMEB-img & 37.3 \\
RESISC45ZeroShot & MIEB & 36.7 \\
RefCOCO & MMEB-img & 35.8 \\
Visual7W-Pointing & MMEB-img & 34.6 \\
N24News & MMEB-img & 34.1 \\
DTDZeroShot & MIEB & 33.1 \\
EuroSATZeroShot & MIEB & 33.0 \\
UCF101ZeroShot & MIEB & 31.9 \\
StanfordCarsZeroShot & MIEB & 30.1 \\
MNISTZeroShot & MIEB & 27.8 \\
WebQAT2ITRetrieval & MIEB & 26.7 \\
FER2013ZeroShot & MIEB & 26.1 \\
CIRR & MMEB-img & 24.8 \\
BirdsnapZeroShot & MIEB & 24.3 \\
MSCOCO & MMEB-img & 23.8 \\
NIGHTSI2IRetrieval & MIEB & 23.3 \\
GTSRBZeroShot & MIEB & 18.2 \\
WebQAT2TRetrieval & MIEB & 17.7 \\
VisDial & MMEB-img & 17.7 \\
CLEVRZeroShot & MIEB & 17.6 \\
CLEVRCountZeroShot & MIEB & 16.2 \\
CIRRIT2IRetrieval & MIEB & 13.6 \\
XFlickr30kCoT2IRetrieval$_{M}$ & MIEB & 10.1 \\
ViDoRe\_infovqa & MMEB-doc & 9.7 \\
VidoreInfoVQARetrieval & MIEB & 9.6 \\
VisRAG\_InfoVQA & MMEB-doc & 9.5 \\
Country211 & MMEB-img & 8.8 \\
Country211ZeroShot & MIEB & 8.6 \\
VisualNewsI2TRetrieval & MIEB & 7.0 \\
OK-VQA & MMEB-img & 6.9 \\
VisualNewsT2IRetrieval & MIEB & 6.4 \\
ViDoRe\_docvqa & MMEB-doc & 3.8 \\
VidoreDocVQARetrieval & MIEB & 3.7 \\
InfographicsVQA & MMEB-img & 3.5 \\
DocVQA & MMEB-img & 2.8 \\
VisRAG\_MP-DocVQA & MMEB-doc & 2.8 \\
VisRAG\_ChartQA & MMEB-doc & 2.1 \\
A-OKVQA & MMEB-img & 1.2 \\
Visual7W & MMEB-img & 1.1 \\
ChartQA & MMEB-img & 0.9 \\
StanfordCarsI2IRetrieval & MIEB & 0.0 \\
\bottomrule
\end{tabular}
\hspace{1.5em}
\begin{tabular}[t]{l l r}
\toprule
\multicolumn{3}{l}{\textbf{Challenge} (46 tasks)} \\
\textbf{Task} & \textbf{Suite} & \textbf{Score} \\
\midrule
VidoreTabfquadRetrieval & MIEB & 5.9 \\
ViDoRe\_tabfquad & MMEB-doc & 5.9 \\
MMLongBench-doc & MMEB-doc & 5.8 \\
OVEN$^{A}$ & MMEB-img & 5.7 \\
Wiki-SS-NQ & MMEB-img & 5.2 \\
VidoreSyntheticDocQAAIRetrieval & MIEB & 4.1 \\
ScienceQA & MMEB-img & 4.1 \\
ViDoRe\_syntheticDocQA\_artificial\_intelligence & MMEB-doc & 4.1 \\
XM3600T2IRetrieval$_{M}$ & MIEB & 4.0 \\
MMLongBench-page & MMEB-doc & 3.6 \\
ViDoRe\_economics\_reports\_v2 & MMEB-doc & 3.5 \\
VidoreSyntheticDocQAEnergyRetrieval & MIEB & 3.3 \\
ViDoRe\_syntheticDocQA\_energy & MMEB-doc & 3.3 \\
ViDoRe\_esg\_reports\_v2 & MMEB-doc & 3.3 \\
ImageNet-A & MMEB-img & 3.0 \\
ViDoRe\_economics\_reports\_v2\_multilingual$_{M}$ & MMEB-doc & 2.8 \\
ViDoRe\_biomedical\_lectures\_v2 & MMEB-doc & 2.8 \\
FashionIQIT2IRetrieval & MIEB & 2.7 \\
OVENIT2ITRetrieval & MIEB & 2.7 \\
VizWizIT2TRetrieval & MIEB & 2.7 \\
Fashion200kI2TRetrieval & MIEB & 2.6 \\
InfoSeekIT2ITRetrieval & MIEB & 2.5 \\
RParisMediumI2IRetrieval & MIEB & 2.5 \\
TextVQA & MMEB-img & 2.4 \\
VisRAG\_ArxivQA & MMEB-doc & 2.4 \\
ROxfordHardI2IRetrieval & MIEB & 2.3 \\
Fashion200kT2IRetrieval & MIEB & 2.1 \\
VidoreSyntheticDocQAHealthcareIndustryRetrieval & MIEB & 1.9 \\
ViDoRe\_syntheticDocQA\_healthcare\_industry & MMEB-doc & 1.9 \\
ViDoRe\_esg\_reports\_human\_labeled\_v2 & MMEB-doc & 1.7 \\
VidoreTatdqaRetrieval & MIEB & 1.6 \\
ViDoRe\_biomedical\_lectures\_v2\_multilingual$_{M}$ & MMEB-doc & 1.6 \\
ViDoRe\_esg\_reports\_v2\_multilingual$_{M}$ & MMEB-doc & 1.6 \\
ViDoRe\_tatdqa & MMEB-doc & 1.6 \\
ViDoRe\_shiftproject & MMEB-doc & 1.5 \\
VidoreShiftProjectRetrieval & MIEB & 1.4 \\
VidoreSyntheticDocQAGovernmentReportsRetrieval & MIEB & 1.3 \\
ViDoRe\_syntheticDocQA\_government\_reports & MMEB-doc & 1.3 \\
VizWiz & MMEB-img & 1.2 \\
VQA2IT2TRetrieval & MIEB & 0.8 \\
OVENIT2TRetrieval & MIEB & 0.4 \\
InfoSeekIT2TRetrieval & MIEB & 0.2 \\
RParisHardI2IRetrieval & MIEB & 0.1 \\
VisRAG\_PlotQA & MMEB-doc & 0.1 \\
CUB200I2IRetrieval & MIEB & 0.0 \\
METI2IRetrieval & MIEB & 0.0 \\
\bottomrule
\end{tabular}
}
\caption{
Tasks in each TRACE group under pool B (A + MobileCLIP2-S0, $\tau$ = 0.064), sorted by the pooled score of the pool's models (Score, 0--100).
Adaptability is fixed by the availability of training data; General and Challenge are split at $\tau$. A superscript $^{A}$ marks the 7 tasks that are in the other non-adaptable group under the paper split (pool A).
A subscript $_{M}$ marks the tasks evaluated over several languages.
}
\label{tab:group_members_B}
\end{table*}

\begin{table*}[p]
\centering
\setlength{\tabcolsep}{3pt}
\scriptsize
\resizebox{\linewidth}{!}{%
\begin{tabular}[t]{l l r}
\toprule
\multicolumn{3}{l}{\textbf{General} (55 tasks)} \\
\textbf{Task} & \textbf{Suite} & \textbf{Score} \\
\midrule
RP2kI2IRetrieval & MIEB & 100.0 \\
SOPI2IRetrieval & MIEB & 100.0 \\
MemotionT2IRetrieval & MIEB & 99.0 \\
MemotionI2TRetrieval & MIEB & 92.7 \\
SugarCrepe & MIEB & 77.9 \\
BLINKIT2IMultiChoice & MIEB & 73.9 \\
TUBerlinT2IRetrieval & MIEB & 69.3 \\
Flickr30kT2IRetrieval & MIEB & 69.0 \\
Flickr30kI2TRetrieval & MIEB & 64.5 \\
AROVisualAttribution & MIEB & 56.6 \\
ImageNet-R & MMEB-img & 51.0 \\
CVBenchDepth & MIEB & 50.2 \\
CVBenchDistance & MIEB & 49.4 \\
AROVisualRelation & MIEB & 48.7 \\
CVBenchRelation & MIEB & 48.6 \\
STS17MultilingualVisualSTS & MIEB & 48.0 \\
VisualSTS17Eng & MIEB & 48.0 \\
GQA & MMEB-img & 45.0 \\
GLDv2I2TRetrieval & MIEB & 39.9 \\
BLINKIT2TMultiChoice & MIEB & 39.4 \\
EDIS & MMEB-img & 39.4 \\
Place365 & MMEB-img & 35.2 \\
AROFlickrOrder & MIEB & 33.9 \\
STS16VisualSTS & MIEB & 33.7 \\
STS12VisualSTS & MIEB & 32.2 \\
BLINKIT2IRetrieval & MIEB & 31.9 \\
SketchyI2IRetrieval & MIEB & 31.7 \\
STS15VisualSTS & MIEB & 30.4 \\
ObjectNet & MMEB-img & 30.4 \\
AROCocoOrder & MIEB & 29.7 \\
CVBenchCount & MIEB & 27.4 \\
FORBI2IRetrieval & MIEB & 22.6 \\
STS13VisualSTS & MIEB & 21.8 \\
ROxfordEasyI2IRetrieval & MIEB & 21.3 \\
ViDoSeek-doc & MMEB-doc & 21.2 \\
STSBenchmarkMultilingualVisualSTS & MIEB & 18.7 \\
VisualSTS-b-Eng & MIEB & 18.7 \\
STS14VisualSTS$^{A}$ & MIEB & 17.7 \\
VisRAG\_SlideVQA$^{A}$ & MMEB-doc & 17.6 \\
ViDoSeek-page$^{A}$ & MMEB-doc & 17.0 \\
BLINKIT2TRetrieval & MIEB & 14.1 \\
WITT2IRetrieval$_{M}$ & MIEB & 13.1 \\
GLDv2I2IRetrieval & MIEB & 12.8 \\
ImageCoDe & MIEB & 12.6 \\
FashionIQ & MMEB-img & 12.4 \\
OVEN & MMEB-img & 12.3 \\
EDIST2ITRetrieval & MIEB & 12.2 \\
Wiki-SS-NQ$^{A}$ & MMEB-img & 11.8 \\
ROxfordMediumI2IRetrieval & MIEB & 11.3 \\
VidoreArxivQARetrieval$^{A}$ & MIEB & 10.6 \\
FGVCAircraftZeroShot$^{A}$ & MIEB & 10.6 \\
ViDoRe\_arxivqa$^{A}$ & MMEB-doc & 10.6 \\
VidoreSyntheticDocQAEnergyRetrieval$^{A}$ & MIEB & 9.3 \\
ViDoRe\_syntheticDocQA\_energy$^{A}$ & MMEB-doc & 9.3 \\
MMLongBench-doc$^{A}$ & MMEB-doc & 9.2 \\
\bottomrule
\end{tabular}
\hspace{1.5em}
\begin{tabular}[t]{l l r}
\toprule
\multicolumn{3}{l}{\textbf{Adaptability} (72 tasks)} \\
\textbf{Task} & \textbf{Suite} & \textbf{Score} \\
\midrule
SciMMIRI2TRetrieval & MIEB & 100.0 \\
SciMMIRT2IRetrieval & MIEB & 100.0 \\
STL10ZeroShot & MIEB & 92.8 \\
ImageNet10Clustering & MIEB & 88.5 \\
HatefulMemesI2TRetrieval & MIEB & 82.9 \\
HatefulMemesT2IRetrieval & MIEB & 82.9 \\
CIFAR10ZeroShot & MIEB & 79.9 \\
Caltech101ZeroShot & MIEB & 75.6 \\
OxfordPetsZeroShot & MIEB & 70.3 \\
CIFAR10Clustering & MIEB & 62.0 \\
Food101ZeroShot & MIEB & 61.1 \\
NIGHTS & MMEB-img & 61.0 \\
VOC2007 & MMEB-img & 60.4 \\
SUN397ZeroShot & MIEB & 59.2 \\
ImageNetDog15Clustering & MIEB & 58.7 \\
WebQA & MMEB-img & 58.2 \\
CIFAR100Clustering & MIEB & 57.8 \\
SUN397 & MMEB-img & 57.4 \\
PatchCamelyonZeroShot & MIEB & 54.3 \\
TinyImageNetClustering & MIEB & 54.2 \\
RenderedSST2 & MIEB & 52.2 \\
MSCOCO\_t2i & MMEB-img & 52.2 \\
HatefulMemes & MMEB-img & 51.2 \\
CIFAR100ZeroShot & MIEB & 50.6 \\
MSCOCO\_i2t & MMEB-img & 50.5 \\
Imagenet1kZeroShot & MIEB & 49.3 \\
RefCOCO-Matching & MMEB-img & 48.7 \\
MSCOCOT2IRetrieval & MIEB & 46.4 \\
VisualNews\_i2t & MMEB-img & 46.1 \\
ImageNet-1K & MMEB-img & 45.6 \\
VisualNews\_t2i & MMEB-img & 43.4 \\
RESISC45ZeroShot & MIEB & 41.5 \\
MSCOCOI2TRetrieval & MIEB & 40.2 \\
StanfordCarsZeroShot & MIEB & 39.8 \\
RefCOCO & MMEB-img & 38.9 \\
Visual7W-Pointing & MMEB-img & 38.2 \\
DTDZeroShot & MIEB & 36.6 \\
UCF101ZeroShot & MIEB & 35.8 \\
EuroSATZeroShot & MIEB & 35.7 \\
N24News & MMEB-img & 34.9 \\
MNISTZeroShot & MIEB & 34.6 \\
FER2013ZeroShot & MIEB & 30.1 \\
BirdsnapZeroShot & MIEB & 30.0 \\
WebQAT2ITRetrieval & MIEB & 29.0 \\
MSCOCO & MMEB-img & 25.6 \\
NIGHTSI2IRetrieval & MIEB & 23.7 \\
GTSRBZeroShot & MIEB & 23.3 \\
CIRR & MMEB-img & 23.2 \\
WebQAT2TRetrieval & MIEB & 19.9 \\
VisDial & MMEB-img & 18.9 \\
CLEVRZeroShot & MIEB & 18.2 \\
CLEVRCountZeroShot & MIEB & 17.1 \\
VisRAG\_InfoVQA & MMEB-doc & 16.6 \\
VidoreInfoVQARetrieval & MIEB & 16.2 \\
ViDoRe\_infovqa & MMEB-doc & 16.2 \\
CIRRIT2IRetrieval & MIEB & 13.7 \\
XFlickr30kCoT2IRetrieval$_{M}$ & MIEB & 12.6 \\
VisualNewsI2TRetrieval & MIEB & 10.9 \\
Country211ZeroShot & MIEB & 10.3 \\
VisualNewsT2IRetrieval & MIEB & 10.2 \\
Country211 & MMEB-img & 10.0 \\
VisRAG\_ChartQA & MMEB-doc & 8.8 \\
OK-VQA & MMEB-img & 7.7 \\
VisRAG\_MP-DocVQA & MMEB-doc & 5.6 \\
VidoreDocVQARetrieval & MIEB & 5.4 \\
ViDoRe\_docvqa & MMEB-doc & 5.4 \\
InfographicsVQA & MMEB-img & 3.7 \\
DocVQA & MMEB-img & 3.2 \\
A-OKVQA & MMEB-img & 1.6 \\
Visual7W & MMEB-img & 1.3 \\
ChartQA & MMEB-img & 1.0 \\
StanfordCarsI2IRetrieval & MIEB & 0.0 \\
\bottomrule
\end{tabular}
\hspace{1.5em}
\begin{tabular}[t]{l l r}
\toprule
\multicolumn{3}{l}{\textbf{Challenge} (44 tasks)} \\
\textbf{Task} & \textbf{Suite} & \textbf{Score} \\
\midrule
VidoreTabfquadRetrieval & MIEB & 8.8 \\
ViDoRe\_tabfquad & MMEB-doc & 8.8 \\
ViDoRe\_syntheticDocQA\_healthcare\_industry & MMEB-doc & 8.1 \\
ImageCoDeT2IRetrieval$^{A}$ & MIEB & 8.0 \\
RParisEasyI2IRetrieval$^{A}$ & MIEB & 8.0 \\
VidoreSyntheticDocQAHealthcareIndustryRetrieval & MIEB & 7.9 \\
VidoreSyntheticDocQAAIRetrieval & MIEB & 7.8 \\
ViDoRe\_syntheticDocQA\_artificial\_intelligence & MMEB-doc & 7.8 \\
ViDoRe\_biomedical\_lectures\_v2 & MMEB-doc & 7.8 \\
Winoground$^{A}$ & MIEB & 6.8 \\
MMLongBench-page & MMEB-doc & 6.4 \\
ViDoRe\_syntheticDocQA\_government\_reports & MMEB-doc & 6.4 \\
VidoreSyntheticDocQAGovernmentReportsRetrieval & MIEB & 6.3 \\
ViDoRe\_economics\_reports\_v2 & MMEB-doc & 6.2 \\
XM3600T2IRetrieval$_{M}$ & MIEB & 5.4 \\
ScienceQA & MMEB-img & 5.1 \\
VisRAG\_ArxivQA & MMEB-doc & 5.1 \\
ImageNet-A & MMEB-img & 4.9 \\
ViDoRe\_esg\_reports\_v2 & MMEB-doc & 4.9 \\
ViDoRe\_biomedical\_lectures\_v2\_multilingual$_{M}$ & MMEB-doc & 4.1 \\
ViDoRe\_economics\_reports\_v2\_multilingual$_{M}$ & MMEB-doc & 4.0 \\
Fashion200kI2TRetrieval & MIEB & 3.9 \\
TextVQA & MMEB-img & 3.8 \\
InfoSeekIT2ITRetrieval & MIEB & 3.5 \\
OVENIT2ITRetrieval & MIEB & 3.5 \\
ViDoRe\_esg\_reports\_human\_labeled\_v2 & MMEB-doc & 3.3 \\
Fashion200kT2IRetrieval & MIEB & 3.1 \\
VizWizIT2TRetrieval & MIEB & 3.1 \\
FashionIQIT2IRetrieval & MIEB & 3.0 \\
VidoreTatdqaRetrieval & MIEB & 3.0 \\
ViDoRe\_tatdqa & MMEB-doc & 2.9 \\
VidoreShiftProjectRetrieval & MIEB & 2.8 \\
ViDoRe\_shiftproject & MMEB-doc & 2.7 \\
RParisMediumI2IRetrieval & MIEB & 2.5 \\
ROxfordHardI2IRetrieval & MIEB & 2.4 \\
ViDoRe\_esg\_reports\_v2\_multilingual$_{M}$ & MMEB-doc & 2.3 \\
VizWiz & MMEB-img & 2.0 \\
VisRAG\_PlotQA & MMEB-doc & 1.0 \\
VQA2IT2TRetrieval & MIEB & 0.9 \\
OVENIT2TRetrieval & MIEB & 0.7 \\
InfoSeekIT2TRetrieval & MIEB & 0.3 \\
RParisHardI2IRetrieval & MIEB & 0.1 \\
CUB200I2IRetrieval & MIEB & 0.0 \\
METI2IRetrieval & MIEB & 0.0 \\
\bottomrule
\end{tabular}
}
\caption{
Tasks in each TRACE group under pool C (B + CLIP ViT-B/16, $\tau$ = 0.090), sorted by the pooled score of the pool's models (Score, 0--100).
Adaptability is fixed by the availability of training data; General and Challenge are split at $\tau$. A superscript $^{A}$ marks the 13 tasks that are in the other non-adaptable group under the paper split (pool A).
A subscript $_{M}$ marks the tasks evaluated over several languages.
}
\label{tab:group_members_C}
\end{table*}

\end{document}